\pdfoutput=1

\documentclass[11pt]{article}
\PassOptionsToPackage{table,dvipsnames}{xcolor}

\usepackage[preprint]{acl}
\usepackage{times}
\usepackage{latexsym}
\definecolor{redbox}{HTML}{FFD6D6}

\usepackage{tcolorbox}
\definecolor{bluebox}{HTML}{FFFFFF}

\usepackage[T1]{fontenc}

\usepackage[utf8]{inputenc}

\usepackage{microtype}

\usepackage{inconsolata}

\newcommand{\quotes}[1]{``#1''}
\usepackage{amsmath}
\usepackage{tikz}

\usepackage[export]{adjustbox}
\usepackage{booktabs}
\usepackage{multirow}
\usepackage{hhline}

\usepackage{algorithm}
\usepackage{algpseudocode}
\usepackage{framed}

\usepackage{hyperref}
\usepackage{arydshln}

\usepackage{amssymb}
\usepackage{pifont}

\usepackage{makecell}

\title{MMCIG: Multimodal Cover Image Generation for Text-only Documents and Its Dataset Construction via Pseudo-labeling}

\author{Hyeyeon Kim$^{1\dagger}$, Sungwoo Han$^{1\dagger}$, $^*$Jingun Kwon$^1$, \\ \textbf{Hidetaka Kamigaito}$^2$, \textbf{and Manabu Okumura}$^3$ \\
 $^1$Chungnam National University, $^2$Nara Institute of Science and Technology (NAIST) \\
 $^3$Institute of Science Tokyo \\
 {\tt \{hyk22,77sungwhan\}@o.cnu.ac.kr} \\
 {\tt jingun.kwon@cnu.ac.kr} \\
 {\tt kamigaito.h@is.naist.jp} \\
 {\tt oku@pi.titech.ac.jp}
 \\}

\begin{document}
\maketitle

\let\thefootnote\relax\footnote{$^*$ Corresponding author.}
\let\thefootnote\relax\footnote{$^\dagger$ Equal contribution.}
\begin{abstract}
In this study, we introduce a novel cover image generation task that produces both a concise summary and a visually corresponding image from a given text-only document. Because no existing datasets are available for this task, we propose a multimodal pseudo-labeling method to construct high-quality datasets at low cost. We first collect documents that contain multiple images with their captions, and their summaries by excluding factually inconsistent instances. Our approach selects one image from the multiple images accompanying the documents. Using the gold summary, we independently rank both the images and their captions. Then, we annotate a pseudo-label for an image when both the image and its corresponding caption are ranked first in their respective rankings. Finally, we remove documents that contain direct image references within texts. Experimental results demonstrate that the proposed multimodal pseudo-labeling method constructs more precise datasets and generates higher quality images than text- and image-only pseudo-labeling methods, which consider captions and images separately. We release our code at: \url{https://github.com/HyeyeeonKim/MMCIG}.

\end{abstract}

\section{Introduction}
Text summarization generates concise summaries by preserving essential information from documents. Given the increasing amount of multimedia content on the web, multimodal summarization (MMS), which produces both a textual summary and selects a relevant image from a document, has attracted significant attention~\cite{10li2018,palaskar-etal-2019-multimodal,liu-etal-2020-multistage,sigirmm,multivideo,zhuang-etal-2024-automatic}.

However, existing MMS methods typically require inputs composed of both text and multiple images, making them unsuitable for text-only scenarios commonly encountered in content creation and media. Additionally, pre-trained image generation models often struggle to produce images closely aligned with textual inputs, necessitating further fine-tuning to improve performance~\cite{lee2023aligningtexttoimagemodelsusing,li2024selma}. Moreover, creating large-scale annotated datasets of document-summary-image pairs for supervised training remains costly and challenging~\cite{zhu-etal-2018-msmo,Zhu_Zhou_Zhang_Li_Zong_Li_2020,jiang-etal-2023-exploiting,Qiu_2024_CVPR}.

To overcome these limitations, we propose a multimodal cover image generation (MMCIG) task that first generates concise summaries and then produces visually aligned images from text-only documents. This task is directly motivated by practical needs such as thumbnail generation for news articles, where creating representative images from summaries can improve content discovery and user engagement~\cite{zhu-etal-2018-msmo,Zhu_Zhou_Zhang_Li_Zong_Li_2020}. To support this, we introduce a multimodal pseudo-labeling method, which is the first systematic approach, to construct high-quality training datasets at low cost: (1) Collect documents with multiple images, their captions, and summaries from the \textit{DailyMail} website.\footnote{\url{https://www.dailymail.co.uk}} (2) Filter documents for factual consistency. (3) Independently rank images and captions by relevance to the gold summaries. (4) Annotate an image with a multimodal pseudo-label when both the image and its corresponding caption are ranked first in their respective rankings, ensuring consistency between the textual and visual content. (5) Remove documents that explicitly reference images in their text.

We compare our multimodal pseudo-labeling method for constructing training datasets with text-only and image-only pseudo-labeling methods. In the text-only pseudo-labeling method, we rely solely on 
caption rankings to annotate a pseudo-label for an image. For the image-only pseudo-labeling method, we use only 
image rankings.

Experimental results demonstrate that our multimodal pseudo-labeling method constructs more precise datasets than the text-only and image-only methods. Furthermore, models fine-tuned on our dataset achieve improved performance in image generation. Human evaluation of both the constructed dataset and model-generated outputs confirms that our multimodal pseudo-labeling method effectively constructs precise datasets at low cost, enabling trained models to generate images closely aligned with summaries.

\section{Multimodal Pseudo-labeling}\label{sec:MMCIG}
The MMCIG task applies to real-world scenarios in which only text inputs are available for user-friendly content that requires both text and images as summaries~\cite{zhu-etal-2018-msmo,Zhu_Zhou_Zhang_Li_Zong_Li_2020}. Due to the absence of suitable datasets, we propose a multimodal pseudo-labeling method to efficiently construct high-quality training datasets. 
Appendix~\ref{appen:overview} provides the detailed dataset construction pipeline.

\noindent \textbf{Filtering for Factual Consistency.}
The \textit{DailyMail} dataset suffers from factual inconsistencies between documents and their summaries; thus, we first filter them using factuality models~\cite{guo-etal-2022-questioning}: $\rm BERTScore_{Art}$, AlignScore, and SummaCscore. Each model evaluates the consistency differently. $\rm BERTScore_{Art}$ computes token-level similarity~\cite{bert-score}, AlignScore measures chunk-sentence alignment~\cite{zha-etal-2023-alignscore}, and SummaCscore considers entailment scores~\cite{Laban2022SummaCRN}. 
We independently remove the lowest-scoring 25\% document-summary pairs and retain only document-summary pairs kept across all models to ensure factuality. For factuality scoring, we directly compare the gold summary with the corresponding document~\cite{guo-etal-2022-questioning}.

\noindent \textbf{Ranking Image and Caption.}
We independently rank the images and their captions from each document using the gold summaries, assuming captions typically provide descriptive information about the images~\cite{jiang-etal-2023-exploiting}. Specifically, we rank images and captions by computing cosine similarity with summaries using CLIP~\cite{radford2021learning} and BERTScore~\cite{bert-score}, respectively.

\noindent \textbf{Annotating Images with Multimodal Consistency.}
By independently ranking the images and captions, we obtain two separate rankings. An image is attached with a multimodal pseudo-label when both the image and its corresponding caption are ranked first in their respective rankings. This method ensures that the pseudo-labeled image is visually relevant to the summary and textually aligned through its caption.

\begin{table}[htbp!]
\renewcommand{\arraystretch}{0.8}
\centering
\resizebox{0.85\columnwidth}{!}{
\begin{tabular}{cccc}
\toprule
\rowcolor{gray!10}
\textbf{    } & \textbf{MMCIG$_{\rm Text}$} & \textbf{MMCIG$_{\rm Image}$} & \textbf{MMCIG$_{\rm Multi}$} \\
\midrule
Train & 140,212 (555.7/55.2) & 140,212 (555.7/55.2) & 48,866 (504.9/55.5) \\
Valid & 4,911 (584.6/55.1) & 4,911 (584.6/55.1) & 1,662 (544.7/55.8) \\
Test & 4,968 (566.8/58.2) & 4,968 (566.8/58.2) & 1,774 (506.0/57.2) \\

\bottomrule
\end{tabular}}
\caption{Statistics of MMCIG datasets. The numbers x/y in parentheses indicate the average document and summary lengths, respectively.  } 
\label{tab:datainfo}
\end{table}

\begin{table}[htbp!]  
\renewcommand{\arraystretch}{0.8}
\centering
\resizebox{0.85\columnwidth}{!}{
\begin{tabular}{cccccccc}
\toprule
\rowcolor{gray!10}
\textbf{Dataset} & \textbf{1} & \textbf{2} & \textbf{3} & \textbf{4} & \textbf{5} & \textbf{6} & \textbf{Avg/Total}\\
\midrule
\multirow{2}{*}{MMCIG$_{\rm Text}$}   & 70.0 & 83.7 & 84.3 & 86.7 & 69.6 & 50 & 77.7 \\
    & (298) & (788) & (779) & (670) & (1,323) & (3) & (3,861) 
\\\cdashline{1-8}\noalign{\vskip 0.5ex}
\multirow{2}{*}{MMCIG$_{\rm Image}$}   & 72.3 & 84.9 & 86.2 & 86.9 & 69.5 & 66.7 & 78.5 \\
   & (306) & (799) & (796) & (672) & (1,322) & (4) & (3,899) 
\\\cdashline{1-8}\noalign{\vskip 0.5ex}
\multirow{2}{*}{MMCIG$_{\rm Multi}$}    &\textbf{86.2} &\textbf{88.2} &\textbf{90.0} &\textbf{88.8} &\textbf{78.0} &\textbf{100}  &\textbf{85.9} \\
   & (250) & (412) & (316) & (207) & (337) & (1) & (1,523) \\

\bottomrule
\end{tabular}}

\caption{Accuracy of MMCIG$_{\rm Text}$, MMCIG$_{\rm Image}$, and MMCIG$_{\rm Multi}$ pseudo-labeling methods on the MSMO test dataset. Columns 1-6 represent the number of gold reference images 
in the documents. The numbers in parentheses represent the number of correctly annotated samples in the human-annotated MSMO test dataset.}
\label{tab:acc}
\end{table}

\begin{table}[t!]
\centering
\renewcommand{\arraystretch}{0.8}
\resizebox{0.45\columnwidth}{!}{%
\begin{tabular}{lcc}
\toprule
\rowcolor{gray!10}
\textbf{Dataset} & \textbf{Alignment} & \textbf{Win} \\
\midrule
Random & \underline{3.60} & 20 \\
MMCIG$_{\rm Multi}$ & \textbf{3.85}$^\dagger$ & \textbf{66} \\
\bottomrule
\end{tabular}}
\vspace{1em}
\caption{Human evaluation results. \quotes{Win} denotes the count of pairwise higher scores. $\dagger$ indicates the improvement is significant (\textit{p}<0.05) compared with the underlined score using paired-bootstrap-resampling with 100,000 random samples~\protect\cite{koehn-2004-statistical}. }
\label{tab:datahumaneval_dataset1}
\end{table}

\noindent \textbf{MMCIG Dataset Statistics.}
Because the \textit{DailyMail} dataset often contains direct references to image information in the documents~\cite{hermann2015teachingmachinesreadcomprehend}, we filter out such documents. Algorithm~\ref{algo:algo1} in the Appendix shows the details. We first split each document into sentences with NLTK, and then tag a POS for each word.\footnote{\url{https://www.nltk.org/}} Next, we build candidate lists for singular nouns (NN$_{\rm word}$: \quotes{photo,} \quotes{image,} \quotes{figure,} \quotes{picture,} \quotes{photograph}) and base-form verbs (VB$_{\rm word}$: \quotes{show,} \quotes{reveal,} \quotes{indicate}). Documents containing sentences with both a tagged noun and verb from these lists are removed. 

We create three versions: MMCIG$_{\rm Text}$ (selecting images via caption ranking only), MMCIG$_{\rm Image}$ (via image ranking only), and MMCIG$_{\rm Multi}$ (via both image and caption rankings).
Table~\ref{tab:datainfo} and Table{~\ref{tab:stat2} in the Appendix show the statistics of the MMCIG datasets. While both MMCIG$_{\rm Text}$ and MMCIG$_{\rm Image}$ share identical dataset sizes due to their reliance on a single modality for labeling, MMCIG$_{\rm Multi}$ contains a smaller number of instances.

\begin{table*}[htbp!]
\begin{adjustbox}{width=0.8\textwidth, center}
    \centering
    \renewcommand{\arraystretch}{0.7}
    \resizebox{\textwidth}{!}{%
    \begin{tabular}{cccccccccc}
    \toprule
    \rowcolor{gray!10}
    & & & & & \multicolumn{2}{c}{\textbf{CLIPScore ($\uparrow$)}} & & & \\
    \cline{6-7}
    \rowcolor{gray!10}
    \multirow{-2}{*}{\textbf{Summary Gen.}} & \multirow{-2}{*}{\textbf{Image Gen.}} & \multirow{-2}{*}{\textbf{Setting}} & \multirow{-2}{*}{\textbf{MMCIG}}  & \multirow{-2}{*}{\textbf{BLIPScore ($\uparrow$)}} & \textbf{Txt-Img} & \textbf{Img-Img}  & \multirow{-2}{*}{\textbf{IQA ($\uparrow$)}}& \multirow{-2}{*}{\textbf{IS ($\uparrow$)}} & \multirow{-2}{*}{\textbf{FID ($\downarrow$)}}\\    
    \midrule
\multirow{11}{*}{Gold} & DALL-E-3 & Pre-trained & - & 20.0 & 26.4 & 52.1 & 0.84  & 9.3 & 84.3 \\\cmidrule(lr){2-10}
&  \multirow{4}{*}{Diffusion-2.1} & Pre-trained & - & 28.8 & 31.5 & 64.2 & 0.97 &  14.8 & \textbf{51.6} \\
&  & \multirow{3}{*}{Fine-tuned} & Text & 29.3 & 31.2 & 65.1 & 0.98 &  \textbf{15.9} & 61.1 \\
&  &  & Image & \underline{29.8} & \underline{31.7} & \underline{66.0} & 0.98 &  15.8 & 56.8 \\
&  &  & Multi & 30.0$^\dagger$ & 32.0$^\dagger$ & 67.1$^\dagger$ & 0.98 &  15.1 & 54.3 \\
&  &  & Image $\rightarrow$  Multi & \textbf{30.1}$^\dagger$ & \textbf{32.1}$^\dagger$ & \textbf{67.2}$^\dagger$ & \textbf{0.99} &  15.1 & 53.4

\\\cmidrule(lr){2-10}
&  \multirow{4}{*}{Dreamlike} & Pre-trained & - & 27.8 & 31.5 & 65.2 & 0.95 & 13.5 & 54.9 \\
&  & \multirow{3}{*}{Fine-tuned} & Text & 28.8 & 30.9 & 64.9 & \textbf{0.98} &  15.5 & 58.6 \\
&  &  & Image & 29.1 & 31.5 & \underline{65.7} & \underline{\textbf{0.98}} &  \textbf{15.6} & 53.7 \\
&  &  & Multi & \textbf{29.2} & 31.4 & 65.7 & \textbf{0.98}$^\dagger$ &  15.4 & \textbf{53.4} \\
&  &  & Image $\rightarrow$ Multi & \textbf{29.2} & \textbf{31.6} & \textbf{66.3}$^\dagger$ & \textbf{0.98} & \textbf{15.6} & 53.5 \\

\midrule

\multirow{11}{*}{\shortstack{Llama-3.2-3B\\-Instruct}} & DALL-E-3 & Pre-trained & - & 20.5 & 26.3 & 51.3 & 0.88  & 9.7 & 79.9 \\\cmidrule(lr){2-10}
&  \multirow{4}{*}{Diffusion-2.1} & Pre-trained & - & 28.5 & 31.7 & 63.9 & 0.97 &  15.6 & \textbf{50.6} \\
&  & \multirow{3}{*}{Fine-tuned} & Text & 29.1 & 31.2 & 64.4 & 0.98 &  15.9 & 62.7 \\
&  &  & Image & \underline{29.8} & \underline{31.9} & \underline{65.4} & \underline{0.98} &  \textbf{16.9} & 55.7 \\
&  &  & Multi & 29.9 & 32.1$^\dagger$ & 66.4$^\dagger$ & 0.98 &  15.6 & 53.8 \\
&  &  & Image $\rightarrow$ Multi & \textbf{30.1}$^\dagger$ & \textbf{32.2}$^\dagger$ & \textbf{66.8}$^\dagger$ & \textbf{0.99}$^\dagger$ &  15.6 & 53.5 \\
\cmidrule(lr){2-10}
&  \multirow{4}{*}{Dreamlike} & Pre-trained & - & 28.0 & 30.2 & 63.9 & 0.95 & 13.6 & 52.9 \\
&  & \multirow{3}{*}{Fine-tuned} &  Text & 28.6 & 31.0 & 64.5 & \textbf{0.98} &  15.6 & 58.1 \\
&  &  & Image & \underline{29.1} & \underline{31.5} & 65.5 & \underline{\textbf{0.98}} &  \textbf{16.4} & 54.9 \\
&  &  & Multi & 29.0 & \textbf{31.6}$^\dagger$ & 64.5 & \textbf{0.98} &  14.5 & \textbf{50.2} \\
&  &  & Image $\rightarrow$  Multi & \textbf{29.2}$^\dagger$ & 31.5 & \textbf{65.7} & \textbf{0.98}$^\dagger$ &  15.9 & 53.9 \\\midrule

\multirow{11}{*}{\shortstack{Qwen2.5-3B\\-Instruct}} & DALL-E-3 & Pre-trained & - & 20.8 & 26.63 & 51.6 & 0.88  & 9.91 & 79.81 \\\cmidrule(lr){2-10}
&  \multirow{4}{*}{Diffusion-2.1} & Pre-trained & - & 28.6 & 31.8 & 63.4 & 0.96 &  14.5 & \textbf{52.4} \\
&  & \multirow{3}{*}{Fine-tuned} & Text & 29.1 & 31.3 & 64.3 & \textbf{0.98} &  15.7 & 61.9 \\
&  &  & Image & \underline{29.7} & \underline{31.8} & \underline{65.4} & \textbf{0.98} &  \textbf{16.4} & 55.8 \\
&  &  &  Multi & 29.7 & 32.0$^\dagger$ & 66.2$^\dagger$ & \textbf{0.98} &  16.3 & 54.3 \\
&  &  & Image $\rightarrow$ Multi & \textbf{30.0}$^\dagger$ & \textbf{32.3}$^\dagger$ & \textbf{66.6}$^\dagger$ & \textbf{0.98} &  15.2 & 52.6 \\
\cmidrule(lr){2-10}
&  \multirow{4}{*}{Dreamlike} & Pre-trained & - & 27.9 & 30.1 & 63.8 & 0.95 & 13.9 & 53.5 \\
&  & \multirow{3}{*}{Fine-tuned} &  Text & 28.5 & 31.0 & 64.3 & \textbf{0.98} &  14.9 & 57.7 \\
&  &  &  Image & \textbf{29.1} & 31.6 & 65.3 & \textbf{0.98} &  \textbf{16.3} & 54.1 \\ 
&  &  &  Multi & \textbf{29.1} & \textbf{31.7} & 64.2 & \textbf{0.98} &  14.8 & \textbf{50.4} \\
&  &  & Image $\rightarrow$ Multi & 29.0 & 31.6 & \textbf{65.4} & \textbf{0.98} &  \textbf{16.3} & 53.4 \\

\bottomrule
\end{tabular}
}
\end{adjustbox}
    \caption{
Experimental results for cover image generation from gold and generated summaries using LLMs. Image $\rightarrow$ Multi indicates that the model was first fine-tuned on MMCIG$_{\rm Image}$ and then further fine-tuned on MMCIG$_{\rm Multi}$. The notations are the same as those in Table~\ref{tab:datahumaneval_dataset1}.
}
    \label{tab:1-2-result}
\end{table*}

\begin{table}[t!]
\centering
\renewcommand{\arraystretch}{0.8}
\resizebox{0.85\columnwidth}{!}{%
\begin{tabular}{ccccc}
\toprule
\rowcolor{gray!10}
\textbf{Model} & \textbf{Setting} & \textbf{MMCIG} & \textbf{Fidelity} & \textbf{Alignment} \\
\midrule
DALL-E-3 & Pre-trained & - & 2.08 & 2.56 \\
\multirow{3}{*}{Diffusion-2.1} & Pre-trained & - & 2.81 & 2.60 \\
 & Fine-tuned & Image & \textbf{2.84} & \underline{2.78} \\
 & Fine-tuned & Image $\rightarrow$ Multi & \textbf{2.84} & \textbf{3.02}$^\dagger$ \\
\bottomrule
\end{tabular}}
\caption{Human evaluation results. The notations are the same as those in Table~\ref{tab:datahumaneval_dataset1}.}
\label{tab:datahumaneval_model2}
\end{table}

\noindent \textbf{MMCIG Dataset Evaluation.}
Training and validation datasets are constructed from our collected data. To evaluate our pseudo-labeling, we construct our test dataset from the MSMO test dataset~\cite{zhu-etal-2018-msmo} since it includes multiple human-annotated gold images for each instance, where each instance consists of a document with multiple images along with its summary. Table~\ref{tab:acc} shows that MMCIG$_{\rm Multi}$ consistently outperforms both MMCIG$_{\rm Text}$ and MMCIG$_{\rm Image}$ by accurately aligning summaries with their corresponding images, even when only one gold reference image is provided among multiple images in a document. Thus, our method effectively constructs high-precision datasets with closely aligned document-summary-image pairs.

We also conducted human evaluation. We sampled 100 images in the MMCIG$\rm _{Multi}$ test dataset and created another dataset (\textbf{Random}) by randomly selecting one gold image from multiple gold images in the MSMO test dataset for the corresponding documents. We used Amazon Mechanical Turk with 80 annotators (US high school or bachelor's degree), who rated the image-summary alignment (1 to 5, 5 is the best). Table~\ref{tab:datahumaneval_dataset1} shows that MMCIG$\rm _{Multi}$ significantly outperformed \textbf{Random}, confirming our method effectively annotates images closely aligned with summaries.

\section{Experiments}
\subsection{Experimental Settings}
\noindent \textbf{Datasets and Implementation Details.}
We used the MMCIG datasets we constructed in Section~\ref{sec:MMCIG} for the cover image generation task. Note that we used only the 1,774 test samples from MMCIG$_{\rm Multi}$ for evaluation in all experiments to ensure data quality and consistency.
To summarize documents, we employed the following LLMs: \texttt{Llama-3.2-3B-Instruct}~\cite{llama32technicalreport} 
and \texttt{Qwen2.5-3B-Instruct}~\cite{qwen2025qwen25technicalreport}.
For image generation, we employed the following models: \texttt{DALL-E-3}~\cite{dalle}, \texttt{stable-diffusion-2-1}~\cite{Rombach_2022_CVPR}, 
and \texttt{dreamlike-photoreal-2.0}~\cite{dream}.
In both summary and image generation, we used greedy decoding and 30 inference steps with a guidance scale of 7.5.

\noindent \textbf{Evaluation Metrics.} 
To assess the generated images, we considered traditional evaluation metrics, including the Fréchet Inception Distance (FID), which evaluates the distance between the probability distributions of gold and generated images~\cite{heusel2018ganstrainedtimescaleupdate}, and the Inception Score (IS), which evaluates the diversity and semantic meaningfulness of generated images~\cite{salimans2016improvedtechniquestraininggans}.
In addition, we used CLIPScore to assess how well the generated images align with both the generated summaries (Txt-Img) and target images (Img-Img)~\cite{hessel-etal-2021-clipscore}.
Furthermore, we employed CLIP Image Quality Assessment (CLIP-IQA) to measure the visual quality of images~\cite{wang2022exploring} and the BLIP score to evaluate how effectively the generated images align with the generated summaries~\cite{li2022blipbootstrappinglanguageimagepretraining}. Appendix~\ref{appen:experi} provides details for experimental settings.

\subsection{Results}

We first evaluate images generated from gold summaries and then evaluate images from generated summaries. Table~\ref{tab:1-2-result} shows the results. While DALL-E-3 suffered difficulties to generate relevant images for summaries, we observed performance gains in BLIP, CLIP and IQA scores with fine-tuned models compared to pre-trained models. Furthermore, MMCIG$_{\rm Multi}$ achieved scores comparable to MMCIG$_{\rm Text}$ and MMCIG$_{\rm Image}$, demonstrating the importance of constructing a high-quality dataset. Fine-tuning first on MMCIG$_{\rm Image}$ and subsequently on MMCIG$_{\rm Multi}$ further improved performance. However, IS and FID scores exhibit inconsistency due to their limited capability in accurately evaluating images produced by recent generation models~\cite{jayasumana2024rethinkingfidbetterevaluation}.

\subsection{Analysis}
\noindent \textbf{Human Evaluation.} We also evaluated gold summaries and generated images by asking annotators to rate image fidelity (reality) and summary-image alignment. Table~\ref{tab:datahumaneval_model2} presents the results. We observed that models fine-tuned on our MMCIG datasets significantly improved alignment between a summary and an image without compromising the fidelity. The relatively lower scores obtained by DALL-E is due to its tendency to generate images in a cartoon-like style. 

\noindent \textbf{Case Study.}
Figure~\ref{fig:casestudy(ours)} shows an example from MMCIG$_{\rm Multi}$. The image and caption contain key information from the summary, highlighted in gray, such as \textit{\quotes{workers injured,} \quotes{last night Photos,} \quotes{in great pain,} and \quotes{clothes burned off.}} This alignment ensures that the selected image is visually relevant to the summary and accurately represents the textual content, which enables us to construct more coherent and relevant document-image-summary pairs.
Appendix~\ref{appen:case} provides further case studies.

\begin{figure}[t]
 \centering
  \includegraphics[width=0.75\columnwidth]{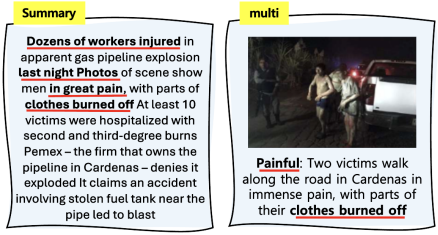}
  \caption{Example case of MMCIG$_{\rm Multi}$.} 
  \label{fig:casestudy(ours)}
\end{figure}

\begin{figure}[t]
 \centering
  \includegraphics[width=0.77\columnwidth]{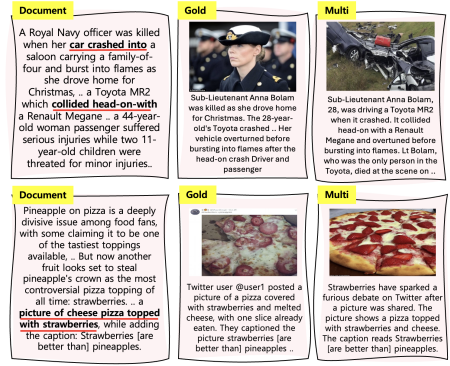}
  \caption{Example of generated images and summaries with their gold references.}
  \label{fig:casestudy(gen)}
\end{figure}
\noindent \textbf{Generated Outputs.} 
Figure~\ref{fig:casestudy(gen)} shows examples of outputs generated by Llama-3.2-3B and Diffusion-2.1 trained on MMCIG$_{\rm Multi}$ compared to gold summaries and images from the test dataset. We observed that the model generates summaries and relevant images. In addition, providing pictorial summaries can enhance user engagement compared to textual summaries alone, which highlights the importance of the proposed task when only textual inputs are available~\cite{zhu-etal-2018-msmo,Zhu_Zhou_Zhang_Li_Zong_Li_2020}.

\section{Conclusion} 
We proposed a novel task of MMCIG that generates textual summaries and their corresponding images from text-only documents to output user-friendly content. Due to the lack of available datasets for this task, we proposed a multimodal pseudo-labeling method to efficiently construct high-quality training datasets. Models trained on our datasets produce informative summaries accompanied by visually corresponding images, as confirmed by automatic and human evaluations.

\section*{Limitations}
Although we proposed the MMCIG task and demonstrated the effectiveness of the proposed multimodal pseudo-labeling method, there are several limitations.

First, constructing MMCIG$_{\rm Multi}$ requires documents with multiple images and their corresponding captions. This may limit the applicability to other domains or languages that lack such rich multimodal datasets. In addition, we primarily considered the dataset from \textit{DailyMail}, which may cause biases related to content style or cultural context due to the nature of this specific domain. However, our approach is the first systematic method for constructing pseudo-labels based on a multimodal approach. In the future, we plan to construct multilingual datasets for different domains for the proposed task.

Second, when generating images for named entities, such as specific people mentioned in the generated summaries, our image generation module struggles to accurately generate images corresponding to these named entities. This may be due to challenges in learning the visual representations for less common entities. In the future, we plan to incorporate external resources to improve image generation for named entities. 

\section*{Ethics Statement}
This section considers the potential ethical issues associated with our model. We proposed MMCIG for the cover image generation task, which is trained on MMCIG$_{\rm Multi}$. MMCIG$_{\rm Multi}$ was constructed from the \textit{DailyMail} dataset, which is a publicly available summarization dataset. Therefore, MMCIG might produce incorrect summaries and images that reflect biases present in the dataset. To mitigate these issues, we cleaned the dataset using factuality models to reduce incorrect or misleading content, since our model generates images based on textual summaries. However, this may not remove all biases present in the dataset. In the future, we plan to consider bias detection methods for better constructing MMCIG$_{\rm Multi}$.

\bibliography{anthology,custom}

\appendix
\section{Multimodal Pseudo-labeling}\label{appen:overview}
\begin{figure*}[ht!]
 \centering
  \includegraphics[width=1.9\columnwidth]{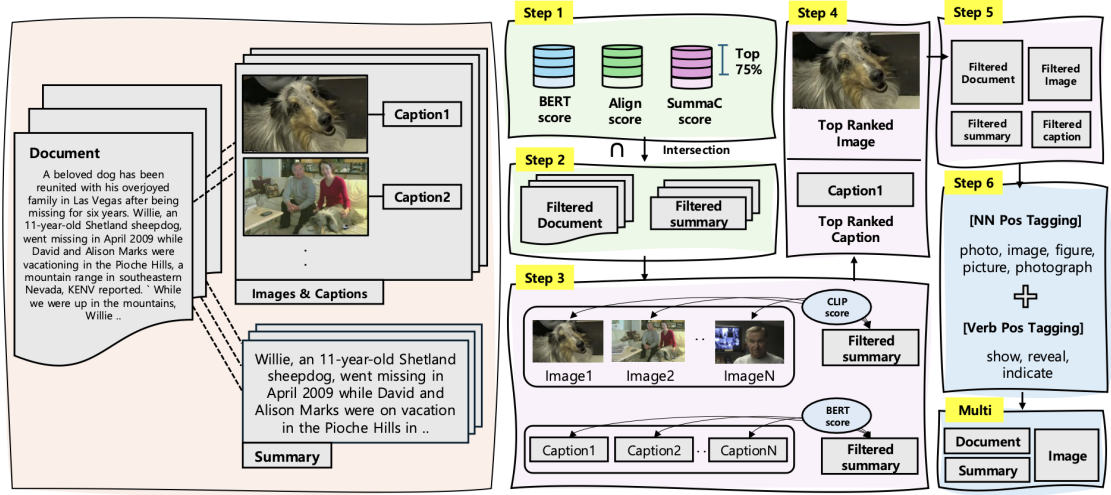}
  \caption{Overview of the MMCIG dataset construction pipeline.}
  \label{fig:datasetconst}
\end{figure*}

\subsection{Overview of Dataset Construction Pipeline}
Figure~\ref{fig:datasetconst} shows an overview of the dataset construction pipeline. We first collect a large-scale multimodal dataset from the \textit{DailyMail} website, containing documents with multiple images, including their captions and summaries. We annotate a pseudo-label by selecting one from the multiple images accompanying each document. Then, we filter out factually inconsistent instances. We independently rank both images and their captions using gold summaries and annotate a multimodal pseudo-label for an image when both the image and its corresponding caption are ranked first in their respective rankings. Finally, we remove instances that contain direct image references within the document.

\subsection{Filtering for Factual Consistency}
We leverage the factuality models $\rm BERTScore_{Art}$, AlignScore, and SummaCscore to automatically evaluate the consistency between documents and gold summaries. $\rm BERTScore_{Art}$ computes token-level similarity~\cite{bert-score} using BERT-based embeddings~\cite{devlin-etal-2019-bert}. AlignScore measures alignment using BERT-based embeddings by chunking the document and splitting the gold summaries into sentences~\cite{zha-etal-2023-alignscore}. SummaCscore segments documents into sentences and uses a natural language inference model to compute entailment scores between sentence pairs by aggregating these scores using \textit{max} and \textit{mean} operators~\cite{Laban2022SummaCRN}.

We remove the lowest 25\% of document-summary pairs based on the scores of each model separately. Finally, we take the intersection of the top of 75\% of document-summary pairs from each model because the different factuality models complement each other in detecting erroneous samples~\cite{guo-etal-2022-questioning}.

\subsection{Filtering for Direct Image Reference}
Algorithm~\ref{algo:algo1} describes how to filter such samples. We begin by applying a sentence splitter from NLTK, which splits a given document into individual sentences. 

Table~\ref{tab:stat2} shows the dataset statistics at each filtering step. Starting from the original dataset, containing 293,966 training, 10,353 validation, and 10,262 test samples, the dataset is successively filtered through stages of factual consistency, multimodal consistency, and POS tagging. After applying factual consistency filtering, the size of the dataset reduces significantly to 140,212 training, 4,911 validation, and 4,968 test instances. Subsequent filtering based on multimodal consistency further reduces the training set to 50,496, validation set to 1,726, and test set to 1,832 samples. Finally, after POS tagging filtering, the dataset comprises 48,866 training, 1,662 validation, and 1,774 test samples, respectively.

For MMCIG$_{\rm Image}$ and MMCIG$_{\rm Text}$, we utilize the dataset obtained after applying factual consistency filtering, resulting in 140,212 training instances, 4,911 validation instances, and 4,968 test instances. For  MMCIG$_{\rm Multi}$, we utilize the dataset after obtained after POS tagging filtering, resulting in 48,866 training instances, 1,662 validation instances, and 1,774 test instances.

\begin{algorithm}[t!]
\caption{Filtering Algorithm.}\label{algo:algo1}
\small
\begin{algorithmic}[1]
\Require~~\\
Documents {\rm{Docs}} = $\{D_1, D_2, \dots, D_n\}$ \\
$\rm NLTK_{ssplit}$ = sentence splitter \\
$\rm NLTK_{tagger}$ = part-of-speech tagger \\
Initialize the filtered documents, {\rm{F}} = [ ] \\
Initialize the {\rm{NN$_{\rm{word}}$}} and {\rm{VB$_{\rm{word}}$}}, respectively
\For{$i=1$ to $n$}
    \State{$S \leftarrow {\rm{NLTK_{ssplit}(D_i)}}$}
    \For{$j=1$ to $|S|$}
        \State{$\rm Tag \leftarrow {\rm{NLTK_{tagger}(S_j)}}$}
        \If{{\rm{Tag$_{\rm{nn}}$}} in {\rm{NN$_{\rm{word}}$}} \textbf{and} {\rm{Tag$_{\rm{vb}}$}} in {\rm{VB$_{\rm{word}}$}}}
            \State{continue}
        \Else{}
            \State{$\rm{F.append(D_i)}$}
        \EndIf
    \EndFor
\EndFor \\
\Return {\rm{F}}
\end{algorithmic}
\end{algorithm}

\begin{table}[ht!]
\renewcommand{\arraystretch}{0.8}
\centering
\resizebox{\columnwidth}{!}{
\begin{tabular}{ccccc}
\toprule
\rowcolor{gray!10}
& \textbf{Original} & \makecell{\textbf{After Factual} \\ \textbf{Consistency}} & \makecell{\textbf{After Multimodal} \\ \textbf{Consistency}} & \makecell{\textbf{After} \\ \textbf{POS tagging}} \\
\midrule
Train & 293,966 & 140,212 & 50,496 & 48,866 \\
Valid & 10,353 & 4,911 & 1,726 & 1,662 \\
Test & 10,262 & 4,968 & 1,832 & 1,774 \\
\bottomrule
\end{tabular}}
\caption{Dataset statistics after each filtering step.}
\label{tab:stat2}
\end{table}

\section{Detailed Experimental Settings}\label{appen:experi}
\subsection{Hyper-parameters}
All experiments were conducted using NVIDIA RTX A6000 GPU.
Table~\ref{tab:hyp-img} shows the hyper-parameters for fine-tuning image generation.

\begin{table}[ht!]
\renewcommand{\arraystretch}{0.8}
\centering
\resizebox{\columnwidth}{!}{
\begin{tabular}{cc}
\toprule
\rowcolor{gray!10}
\multicolumn{2}{c}{\textbf{Hyperparameters}} \\
\midrule
seed & {42}  \\
\midrule
number of training epochs &{20}  \\
\midrule
early stopping &{3}  \\
\midrule
batch size &{20}  \\
\midrule
image resolution &{768}  \\
\midrule
optimizer &{AdamW}  \\
\midrule
learning rate &{3e-7}  \\
\midrule
learning rate scheduler &{constant}  \\

\bottomrule
\end{tabular}}
\caption{Hyper-parameters for image generation models.}
\label{tab:hyp-img}
\end{table}

Table~\ref{tab:hyp-sum} shows the hyper-parameters for fine-tuning summary generation.
We incorporated a parameter-efficient fine-tuning method~\cite{peft}, specifically low-rank adapters, which combine trainable low-rank matrices with the frozen weights in transformer and diffusion layers~\cite{hu2022lora}. We adapted only specific layers to optimize training efficiency: the query, key, value, output layers in LLMs. 

\begin{table}[ht!]
\renewcommand{\arraystretch}{0.8}
\centering
\resizebox{\columnwidth}{!}{
\begin{tabular}{cc}
\toprule
\rowcolor{gray!10}
\multicolumn{2}{c}{\textbf{Hyperparameters}} \\
\midrule
seed & {42}  \\
\midrule
number of training epochs &{20}  \\
\midrule
early stopping &{3}  \\
\midrule
batch size &{8}  \\
\midrule
optimizer &{AdamW}  \\
\midrule
learning rate &{1e-4}  \\
\midrule
lora rank & {8} \\
\midrule
lora alpha & {16} \\
\midrule
lora dropout & {0.1} \\
\midrule
target modules & {query, key, value, and output} \\
\bottomrule
\end{tabular}}
\caption{Hyper-parameters for text generation models.}
\label{tab:hyp-sum}
\end{table}

\subsection{Detailed Evaluation Metrics}

\noindent \textbf{CLIPScore (Img-Txt)}~\cite{hessel2022clipscorereferencefreeevaluationmetric} evaluates the alignment between generated images and their textual descriptions. It computes the cosine similarity between image and text embeddings obtained using the CLIP model~\cite{radford2021learning}. A higher CLIPScore denotes greater semantic consistency between the image and text.

\noindent \textbf{CLIPScore (Img-Img)} evaluates the alignment between generated images and their gold images. It computes the cosine similarity between two image embeddings obtained using the CLIP model. A higher CLIPScore denotes greater semantic consistency between two images.

\noindent \textbf{BLIPScore (Img-Txt)} evaluates the alignment between generated images and their textual descriptions. It computes the cosine similarity between image and text embeddings obtained using the BLIP model~\cite{li2022blipbootstrappinglanguageimagepretraining}.
A higher BLIPScore denotes greater semantic consistency between the image and text.

\noindent \textbf{CLIP-IQA}~\cite{wang2022exploring} calculates image quality scores by computing cosine similarities between image embeddings and textual embeddings derived from paired antonym prompts using the CLIP model. A higher IQA score with positive prompts denotes better perceived quality.

\subsection{DALL-E-3 Prompt for Image Generation}
We used the OpenAI API to generate relevant images with DALL-E-3. Although we experimented with various prompts to enhance performance, the results did not significantly differ.
\begin{tcolorbox}[title=Prompt for Image Generation (\textsc{DALL-E-3}), boxrule=1pt, colback=white, coltitle=white, colframe=blue!50, colbacktitle=blue!60]
\small
\texttt{Please generate a relevant image.} \\ 
\texttt{<Summary>}
\end{tcolorbox}
\quotes{<Summary>} indicates the placeholder for a summary.

\subsection{Document Summarization}
For n-gram matching automatic evaluation metrics, we used ROUGE-1 (R-1), -2 (R-2), and -L (R-L) to assess summarization performance. We also considered BERTScore to assess the contextual similarity between generated and gold summaries. We fine-tuned models on MMCIG$_{\rm Multi}$.
Table~\ref{tab:summarization} shows the results.

\begin{table}[htbp!]  
\renewcommand{\arraystretch}{0.8}
\centering
\resizebox{\columnwidth}{!}{
\begin{tabular}{cccccc}
\toprule
\rowcolor{gray!10}
\textbf{Model} & \textbf{Setting} & \textbf{R-1} & \textbf{R-2} & \textbf{R-L} & \textbf{BS} \\
\midrule
\multirow{2}{*}{Llama-3B}   & Pre-trained & 37.3 & 14.9 & 23.7 & 41.7 \\
                            & Fine-tuned & 47.8 & 24.7 & 34.3 & 48.8 \\\midrule

\multirow{2}{*}{Qwen-3B}    & Pre-trained & 35.0 & 13.0 & 21.9 & 40.3 \\
                            & Fine-tuned & 45.7 & 22.8 & 32.7 & 47.4 \\

\bottomrule
\end{tabular}}

\caption{Experimental results of document summarization.}
\label{tab:summarization}
\end{table}

\section{Additional Case study}\label{appen:case}
\noindent \textbf{Limitations of Text-only and Image-only Pseudo-labeling.}
Figure~\ref{fig:casestudy(mis_1)} presents examples from MMCIG$_{\rm Text}$ and MMCIG$_{\rm Image}$. In MMCIG$_{\rm Image}$, images are selected solely based on image information, whereas in MMCIG$_{\rm Text}$, captions are considered to form document-image-summary pairs. However, ranked images and captions do not always align; thus, the constructed image-summary pairs often lack coherence and relevance~\cite{Zhu_Zhou_Zhang_Li_Zong_Li_2020}.  

In the first example, the summary relies on textual information, while the selected image does not accurately capture entities such as person, \textit{\quotes{Michael Simon de Normier.}} 
In the second example, the key information \textit{\quotes{The files were being stored}} is clearly represented in the first ranked image in MMCIG$_{\rm Image}$, while the MMCIG$_{\rm Text}$ image fails to capture this. This demonstrates the limitation of relying exclusively on either caption or image information.

\begin{figure}[t]
 \centering
  \includegraphics[width=0.95\columnwidth]{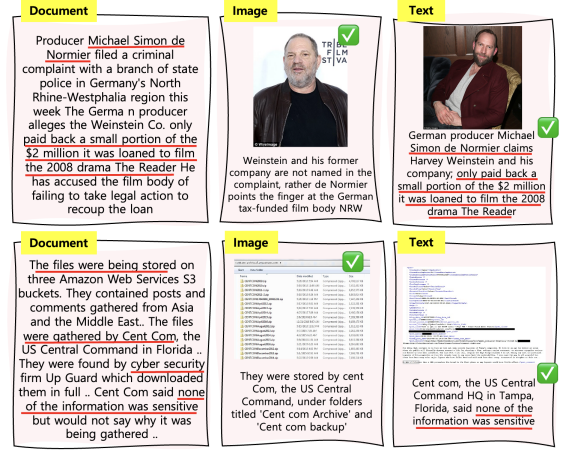}
  \caption{Two example cases of MMCIG$_{\rm Text}$ and MMCIG$_{\rm Image}$. Check mark indicates the first image or caption based on rankings.} 
  \label{fig:casestudy(mis_1)}
\end{figure}

\noindent \textbf{Generated Outputs.} Figure~\ref{fig:all} shows example outputs generated by the stable-diffusion-2.1 models based on generated summaries from fine-tuned Llama-3B-Instruct. DALL-E-3 tended to produce cartoon-like images, whereas diffusion models generated more realistic images. Additionally, the model trained on MMCIG$_{\rm Multi}$ produced images closely aligned with their corresponding summaries.

\begin{figure*}[t]
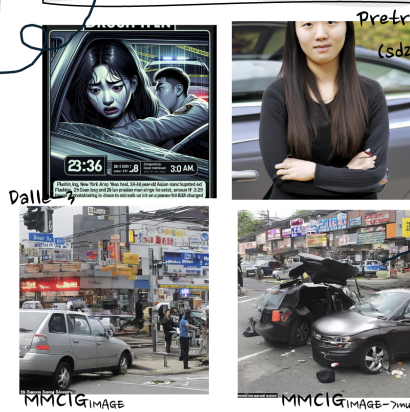

\centering
    \includegraphics[width=1.8\columnwidth]{case_addi.pdf}
    \includegraphics[width=1.8\columnwidth]{case_addi2.pdf}
    \caption{Another example of generated images and summaries. } 
\label{fig:all}
\end{figure*}
\end{document}